\documentclass[a4paper,fleqn]{cas-dc}

\usepackage[numbers]{natbib}
\usepackage{hyperref}
\def\tsc#1{\csdef{#1}{\textsc{\lowercase{#1}}\xspace}}
\tsc{WGM}
\tsc{QE}
\tsc{EP}
\tsc{PMS}
\tsc{BEC}
\tsc{DE}
\makeatletter
\renewcommand{\printorcid}{} 
\makeatother
\shorttitle{}
\begin{document}
\let\WriteBookmarks\relax
\def\floatpagepagefraction{1}
\def\textpagefraction{.001}
\shortauthors{Peilin Zhang et~al.}

\title [mode = title]{TABNet: A Triplet Augmentation Self-Recovery Framework with Boundary-Aware Pseudo-Labels for Medical Image Segmentation}



\author[1]{Peilin Zhang}
\fnmark[1]
\affiliation[1]{
    organization={School of Information Science and Technology, Northwest University}, 
    city={Xi’an}, 
    country={China}
}
\author[1]{Shaouxan Wu}
\fnmark[1]
\author[1]{Jun Feng}
\author[1]{Zhuo Jin}
\author[1]{Zhizezhang Gao}
\author[2]{Jingkun Chen}
\author[1]{Yaqiong Xing}
\cormark[1]
\ead{xkpg@nwu.edu.cn}
\author[1]{Xiao Zhang}
\cormark[1]
\ead{xiaozhang@nwu.edu.cn}
\affiliation[2]{
    organization={Institute of Biomedical Engineering, Department of Engineering Science, University of Oxford}, 
    city={Oxford}, 
    country={UK}
}

\cortext[cor1]{Corresponding author}
\fntext[fn1]{Co-first author.}

\begin{abstract}
\noindent \textit{Background and objective}: Medical image segmentation is a core task in various clinical applications. However, acquiring large-scale, fully annotated medical image datasets is both time-consuming and costly. Scribble annotations, as a form of sparse labeling, provide an efficient and cost-effective alternative for medical image segmentation. However, the sparsity of scribble annotations limits the feature learning of the target region and lacks sufficient boundary supervision, which poses significant challenges for training segmentation networks. 

\noindent \textit{Methods}: We propose TABNet, a novel weakly-supervised medical image segmentation framework, consisting of two key components: the triplet augmentation self-recovery (TAS) module and the boundary-aware pseudo-label supervision (BAP) module. The TAS module enhances feature learning through three complementary augmentation strategies: intensity transformation improves the model's sensitivity to texture and contrast variations, cutout forces the network to capture local anatomical structures by masking key regions, and jigsaw augmentation strengthens the modeling of global anatomical layout by disrupting spatial continuity. By guiding the network to recover complete masks from diverse augmented inputs, TAS promotes a deeper semantic understanding of medical images under sparse supervision. The BAP module enhances pseudo-supervision accuracy and boundary modeling by fusing dual-branch predictions into a loss-weighted pseudo-label and introducing a boundary-aware loss for fine-grained contour refinement. 

\noindent \textit{Results}: Experimental evaluations on two public datasets, ACDC and MSCMRseg, demonstrate that TABNet significantly outperforms state-of-the-art methods for scribble-based weakly supervised segmentation. Moreover, it achieves performance comparable to that of fully supervised methods. 

\noindent \textit{Conclusions}: TABNet effectively addresses the limitations of sparse scribble annotations by integrating multi-view augmentation and boundary-aware supervision. It offers a promising and practical solution for accurate medical image segmentation under weak supervision. The code is available at \url{https://github.com/IPMI-NWU/TABNet}. 
\end{abstract}

\maketitle

\section{Introduction}

Fully annotated medical image datasets are difficult to obtain and require experienced clinicians to provide comprehensive annotations, making the process both time-consuming and labor-intensive. To alleviate the annotation burden, researchers have developed weakly supervised segmentation methods \cite{AIL,10909610}. Among these, scribble annotations have emerged as a prevalent solution owing to their operational flexibility and annotation efficiency, enabling annotators to delineate target regions through simple line drawings \cite{Embracing}. This methodology substantially reduces both temporal expenditure and manual labor compared to exhaustive pixel-wise annotation. However, scribble annotations present several challenges: (1) The sparsity of scribble annotations limits the model’s ability to learn complete region semantics , which often leads to over-segmentation or under-segmentation and ultimately degrades overall segmentation performance. (2) the lack of supervision of target boundaries often leads to segmentation errors at object edges. 

Various deep learning-based methods have been proposed to address these challenges \cite{wang2025deep,MICCAI,zhang2024anatomy}, primarily by leveraging data augmentation or pseudo-labeling techniques to enhance feature learning and supervision signals. CycleMix \cite{CycleMix} leverages the spatial perturbation mechanism of PuzzleMix to generate mixed images and corresponding labels, enhancing the model’s ability to capture structural information and enforcing consistency in pseudo-label learning. Additionally, DMPLS \cite{DMPLS} generates pseudo-labels by randomly mixing two prediction results, guiding the model to learn discriminative features across predictions and improving training stability and pseudo-label quality under weak supervision. Despite these efforts, existing methods still face notable limitations. While various weakly supervised methods have been proposed, they still face notable limitations. Most existing perturbation strategies rely on single-type or simplistic combinations of augmentations, limiting their ability to capture the diverse semantics in medical images. Moreover, pseudo-labels are often generated by random weighting or averaging, without explicit boundary modeling, leading to suboptimal supervision and reduced segmentation accuracy. 

To comprehensively learn multi-dimensional features under sparse supervision, we propose a TAS module. This module integrates three complementary image augmentation strategies: intensity perturbation simulates grayscale variations to encourage the network to learn texture-level features; cutout augmentation occludes organ regions, forcing the model to focus on local anatomical structures; jigsaw augmentation disrupts spatial continuity, guiding the model to understand global spatial layouts and organ configurations. These three augmentations are applied in parallel branches, enabling the model to capture texture, local structure, and global context separately. In addition, a self-recovery mechanism is introduced, requiring the model to reconstruct complete segmentation masks from heavily augmented inputs, thereby further enhancing its ability to represent semantic structures. Compared to conventional single-path augmentation methods, TAS significantly broadens the model's representational capacity by introducing diverse perturbations from multiple perspectives, resulting in more robust and comprehensive feature learning\cite{chen2024dynamic,CHEN2025103385}. 

To address the challenges of low-quality pseudo-labels and imprecise boundary predictions, we propose a BAP module. Unlike previous methods that rely on random or fixed weighting to construct pseudo-labels, BAP adopts a supervision-loss-guided adaptive fusion mechanism. Specifically, predictions from two decoder branches are weighted based on their pixel-level cross-entropy losses, enabling the dynamic generation of high-confidence pseudo-labels and improving the accuracy of the supervision signal. In addition, we introduce a boundary-aware loss function that explicitly constrains shape consistency at object boundaries. This encourages the model to learn more precise and coherent contours, particularly for complex anatomical structures with ambiguous or fine-grained boundaries. 

The main contributions of this paper are as follows: 1) We propose TABNet, a weakly supervised medical image segmentation framework based on scribble annotations. TABNet integrates a TAS module and a BAP module to achieve accurate segmentation under sparse supervision. Our framework outperforms state-of-the-art methods on two public datasets; 2) We develop the TAS module, which employs intensity, cutout, and jigsaw augmentations to guide the model in learning texture, local, and global structural features, respectively. Combined with a self-recovery strategy, TAS significantly enhances the model’s segmentation capability under limited supervision. 3) We design the BAP module to dynamically generate high-quality pseudo-labels with loss-weighted strategies and refine boundary supervision, enhancing segmentation performance, and optimizing boundary delineation. 

\section{Related Works}
\subsection{Data Augmentation in Scribble-Supervised Segmentation}

Data augmentation is a crucial strategy for enhancing model robustness and generalization, and has been extensively studied in scribble-supervised medical image segmentation \cite{HE2024102984,HUANG2025111416,10927851}. The main idea is to guide the model to learn invariant and transferable representations by presenting it with diverse variants of the same input. 

For instance, Bortsova et al. \cite{BortsovaConsistency} proposed a dual-branch network that takes differently augmented versions of the same image and enforces consistency between their segmentation outputs, improving model stability. Li et al. \cite{Transconsistent} introduced transformation consistency, aligning predictions on original and augmented images (e.g., rotations, flips) to enhance generalization. 
Zhang et al. \cite{CycleMix} applied a hybrid augmentation strategy with cycle-consistency, enforcing agreement across multiple augmentations to improve segmentation performance. Zhang et al. \cite{ShapePU} proposed Cutout-based regularization, requiring the model to maintain consistent predictions even when parts of the image are occluded, thereby learning robust shape representations. 

In contrast to these methods, we propose a self-restoration augmentation strategy that combines intensity augmentation, jigsaw augmentation, and cutout augmentation. Specifically, intensity perturbation simulates grayscale variations to encourage the network to learn texture-level features; cutout augmentation occludes organ regions, forcing the model to focus on local anatomical structures; jigsaw augmentation disrupts spatial continuity, guiding the model to understand global spatial layouts and organ configurations. Based on these augmentations, we further design a self-restoration mechanism, where features and context learned from the first two augmentations are used to guide reconstruction of the cropped organ regions. This design significantly improves the model’s ability to perceive and restore target structures under sparse supervision. 

\subsection{Pseudo Labeling in Scribble-Supervised Segmentation}

Pseudo labels have been widely adopted in scribble-supervised image segmentation tasks as an effective means to compensate for the limited supervision provided by sparse annotations \cite{Lei,DMPLS}. The core idea is to leverage the model's own predictions to generate supervisory signals for unlabeled regions. To improve the quality of these pseudo labels, many recent methods utilize multi-branch architectures or model ensembles, thereby enhancing the model’s generalization capability. 

For instance, Luo et al. \cite{DMPLS} proposed dynamically merging predictions from two decoders to increase pseudo label diversity. Wu et al. \cite{CompetetoWin} selected high-confidence regions based on confidence maps from multiple branches. Han et al. \cite{Han} employed a three-branch network with varying dilation rates and averaged feature responses under perturbations to construct pseudo labels. Yang et al. \cite{PacingPseudo} introduced a dual-branch architecture with cross-pseudo supervision to enforce consistency between branches. Zhang et al. \cite{HidesClass} incorporated class activation maps and scribble-derived semantic cues for pseudo label generation. These methods predominantly rely on confidence estimation or feature consistency to guide pseudo label fusion. In terms of fusion strategy, they commonly adopt random selection \cite{DMPLS}, which may not always reflect the true reliability of each branch. 

In contrast to prior methods that generate pseudo-labels via random weighting or averaging, we propose a loss-guided pseudo-label fusion strategy that dynamically weighs predictions based on their supervised loss within labeled regions, thereby improving pseudo-label reliability. To further enhance boundary quality, we introduce a boundary-aware loss that enforces consistency between fused pseudo-labels and branch predictions, particularly along anatomical boundaries. Together, these strategies improve the accuracy and 
stability of pseudo-labels while enhancing boundary segmentation under sparse supervision. 

\begin{figure*}[tbp] 
  \centering
  \includegraphics[width=\textwidth]{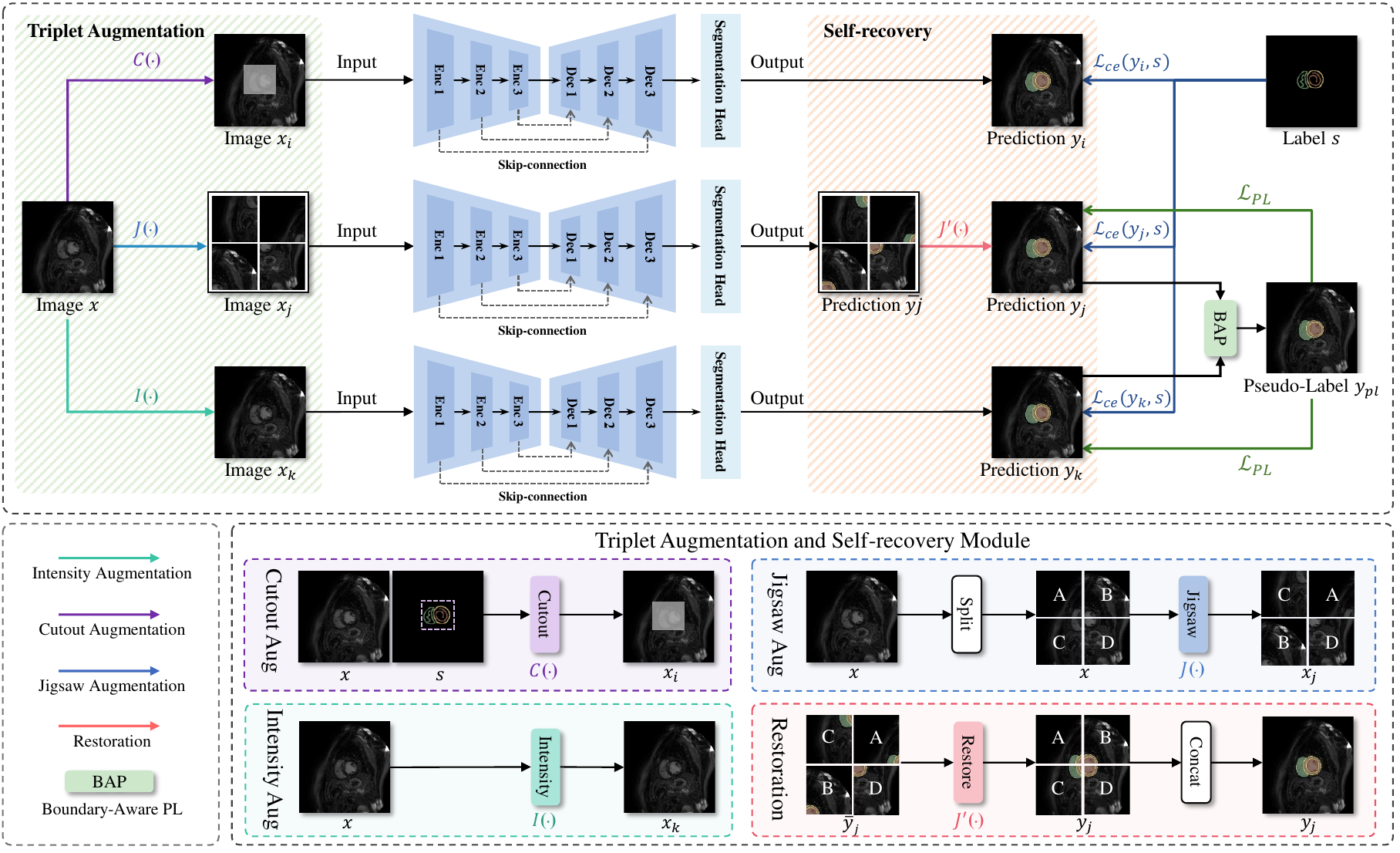} 
  \caption{Overview of the proposed TABNet framework. The model combines triplet augmentation and self-recovery to produce multi-scale predictions, and improves PL quality and boundary accuracy via loss-weighted and boundary constraints. }
  \label{fig:fig0}
\end{figure*}

\section{Method}
As shown in Fig.\ref{fig:fig0}, our scribble-guided medical image segmentation framework consists of two main modules. First, the TAS module generates three distinct augmented versions of the input image and feeds them into a shared-weight segmentation network, which then leverages a self-recovery strategy to encourage the model to focus on learning the target region features. Second, the BAP module integrates predictions from intensity and jigsaw augmentation branches, dynamically generates pseudo-labels with loss-based weighting, and refines target boundaries to enhance segmentation performance. 

\subsection{Triple Augmentation Self-Recovery}
In scribble-supervised segmentation, the limited annotation only marks a few pixel locations of the target organ, which poses a significant challenge to learning complete and discriminative representations\cite{HAN2024103274,10.1145/3581783.3612056,XU2024107744}. To address this issue, we introduce the TAS module,  designed to enhance feature learning by encouraging the model to reconstruct the full organ structure from incomplete or perturbed observations. This module incorporates three complementary data augmentation strategies: intensity perturbation enhances the model’s perception of texture variations; cutout augmentation encourages the learning of local anatomical structures by masking organ regions; jigsaw augmentation strengthens the modeling of global anatomical layout by disrupting spatial continuity. 

Concretely, given an input image $x$, we generate three augmented versions: 
\begin{equation}
x_i = C(x), ~~ x_j = J(x), ~~ x_k = I(x),
\end{equation}
where $C(\cdot)$ denotes cutout augmentation, which masks the foreground region by referring to the maximum bounding box inferred from the scribble annotation. This forces the model to recover semantic features from the occluded core region, thereby enhancing its understanding of local anatomical structures. $J(\cdot)$ refers to jigsaw augmentation, which randomly shuffles non-overlapping patches of the input image, leading to disrupted spatial structure. This enhances the model’s capacity to capture and model global anatomical structures. $I(\cdot)$ denotes intensity augmentation, which applies random brightness and contrast shifts to improve the model’s capacity to perceive and learn texture patterns specific to medical imaging. 

Each augmented image is then independently passed through the segmentation network, producing the predictions $y_i$, $\bar{y_j}$, and $y_k$. However, since the jigsaw transformation disturbs the spatial layout, the resulting prediction $\bar{y_j}$ is misaligned with the original annotations. To address this, we apply an inverse jigsaw operation $J'(\cdot)$ that reorders the prediction map back to the original configuration: 

\begin{equation}
y_j = J'(\bar{y_j}).
\end{equation}

The model is trained by applying a cross-entropy loss on each augmented prediction, computed only over the annotated pixels from the sparse scribble labels: 

\begin{equation}
\mathcal{L}_{ce}(y, s) = \sum_{i\in \Omega_s}^{} \sum_{k\in K}^{}-s^k_i~\log~y^k_i ,
\end{equation}

\begin{equation}
\mathcal{L}_{TAS} = \mathcal{L}_{ce}(y_i, s) + \mathcal{L}_{ce}(y_j, s) + \mathcal{L}_{ce}(y_k, s),
\end{equation}
where $s$ denotes the scribble annotation, $K$ is the number of categories, and $\Omega_s$ represents the pixel coordinates labeled in $s$. 

By simultaneously training the model to predict consistent outputs under occlusion, spatial disorder, and intensity shifts, TAS promotes a more robust and generalizable feature representation. In particular, intensity perturbation simulates grayscale variations to encourage the network to learn texture-level features; cutout augmentation occludes organ regions, forcing the model to focus on local anatomical structures; jigsaw augmentation disrupts spatial continuity, guiding the model to understand global spatial layouts and organ configurations. Together, these augmentations complement each other and help the network to generalize beyond the sparse supervision, ultimately improving segmentation performance under weak annotation settings. 

\subsection{Boundary-Aware Pseudo-Label Supervision}

Under scribble supervision, due to the sparse annotations and lack of boundary information, the model tends to produce inaccurate predictions with incomplete target regions and blurry boundaries. To address this issue, we propose a BAP Module, which aims to generate more reliable pseudo-labels by fusing multi-branch predictions and introducing boundary supervision to guide the model in learning precise object contours\cite{SU2024103111,ZHOU2024107777}. 

The overall structure of this module is illustrated in Fig.\ref{fig:fig1}, which demonstrates the detailed process of pseudo-label fusion and boundary supervision. We first obtain predictions $y_j$ and $y_k$ from the jigsaw augmentation branch and intensity augmentation branch, respectively. The dynamic weights are computed based on the cross-entropy loss with respect to the scribble annotations: 

\begin{figure}[tbp] 
  \centering
  \includegraphics[width=1\columnwidth]{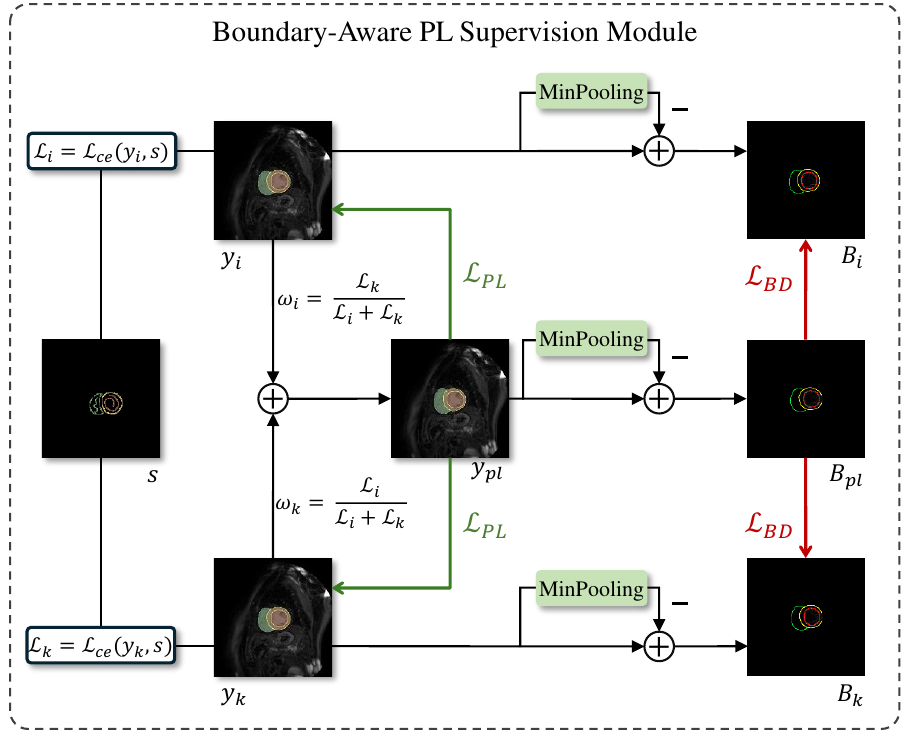} 
  \caption{Boundary-aware pseudo-label supervision enhances segmentation by combining loss-weighted fusion with soft boundary extraction. }
  \label{fig:fig1}
\end{figure}

\begin{equation}
    w_j, w_k\!=\!\frac{\mathcal{L}_{ce}(y_k,\! s)}{ \mathcal{L}_{ce}(y_j,\! s)\!+\!\mathcal{L}_{ce}(y_k, \!s)}, ~~\frac{\mathcal{L}_{ce}(y_j,\! s)}{ \mathcal{L}_{ce}(y_j, \!s)\!+\!\mathcal{L}_{ce}(y_k,\! s)}\!,
\end{equation}
where $s$ denotes the scribble annotations and $\mathcal{L}_{ce}$ is the cross-entropy loss. This weighting strategy suppresses the influence of low-quality predictions, thereby enhancing the reliability and representativeness of the fused pseudo-labels. 
The pseudo-label $y_{pl}$ is then generated by weighted fusion and class-wise maximum selection: 

\begin{equation}
    y_{pl} = argmax([w_j, w_k] \cdot [y_j, y_k]^T).
\end{equation}

To promote prediction consistency, we apply Dice loss to measure the similarity between $y_{pl}$ and each individual prediction, $\Omega_y = \left\{y_j, y_k\right\}$: 

\begin{equation}
    \mathcal{L}_{Dice}(y, y_{pl}) = 1 - \frac{2|y \cap y_{pl}|}{|y| + |y_{pl}|},
\end{equation}
\begin{equation}
    \mathcal{L}_{PL} = \sum_{y \in \Omega_{y}} \mathcal{L}_{Dice}(y, y_{pl}).
\end{equation}

Although the fused pseudo-labels provide stronger region-level information, they may still lack boundary precision. To further refine the boundary predictions, we introduce a boundary-aware supervision mechanism, which extracts boundary maps from $y_j$, $y_k$, and $y_{pl}$ to constrain the outputs of different branches. Specifically, the boundary map is computed as the difference between the original prediction and its eroded version through a soft erosion operation. The boundary-aware loss $\mathcal{L}_{BD}$ is defined as the Dice loss between boundary maps, encouraging consistency between each branch and the pseudo-label: 

\begin{equation}
    B = ReLU(y-{MinPooling}(y)),
\end{equation}

\begin{equation}
    \mathcal{L}_{BD} = \sum_{B \in \Omega_B}(1 - \frac{2 \cdot \sum (B \cdot B_{pl}) + \epsilon}{\sum B + \sum B_{pl} + \epsilon}),
\end{equation}
where $MinPooling(\cdot)$ performs a soft erosion operation on the prediction results, and the boundary map is obtained by subtracting the eroded image from the original image, and $\epsilon$ is $10^{-5}$, $\Omega_B = \left\{B_j, B_k\right\}$ refers to the extraction of the boundaries from two prediction results, $B_{pl}$ represents the boundary extracted from the pseudo-label.The boundary-aware pseudo-label supervision allows the model to focus more effectively on the boundary regions, resulting in clearer and more consistent boundary predictions. 

The final optimization objective is as follows: 
\begin{equation}
    \mathcal{L} =\lambda_1 \mathcal{L}_{TAS} + \lambda_2 \mathcal{L}_{PL} + \lambda_3\mathcal{L}_{BD}
    \label{eq_11}.
\end{equation}

\begin{table*}[tbp]
  \centering
  \small
  \caption{The Dice score of our method on the MSCMRseg and ACDC datasets is compared with various other methods. Bold indicates the best performance among weakly supervised methods. }
  \renewcommand\arraystretch{1.2}
  \setlength{\tabcolsep}{10pt}
  \begingroup
  \fontfamily{ptm}\selectfont
  \begin{tabular}{l|cccc|cccc}
  \toprule
  \multirow{2}{*}{Method} & \multicolumn{4}{c|}{MSCMRseg} & \multicolumn{4}{c}{ACDC} \\
  \cmidrule{2-9}
   & LV & MYO & RV & Avg & LV & MYO & RV & Avg \\
  \midrule
  \textcolor{blue}{\scalebox{0.8}{$\spadesuit$}}~UNet\scalebox{0.8}{+}~\cite{CycleMix} & $.494\scalebox{0.75}{$\pm$}$\scalebox{0.75}{.08} & $.583\scalebox{0.75}{$\pm$}$\scalebox{0.75}{.06} & $.057\scalebox{0.75}{$\pm$}$\scalebox{0.75}{.02} & $.378$ & $.785\scalebox{0.75}{$\pm$}$\scalebox{0.75}{.20} & $.725\scalebox{0.75}{$\pm$}$\scalebox{0.75}{.15} & $.746\scalebox{0.75}{$\pm$}$\scalebox{0.75}{.20} & $.752$ \\
  \textcolor{blue}{\scalebox{0.8}{$\spadesuit$}}~Puzzle Mix~\cite{PuzzleMix} & $.061\scalebox{0.75}{$\pm$}$\scalebox{0.75}{.02} & $.634\scalebox{0.75}{$\pm$}$\scalebox{0.75}{.08} & $.028\scalebox{0.75}{$\pm$}$\scalebox{0.75}{.01} & $.241$ & $.663\scalebox{0.75}{$\pm$}$\scalebox{0.75}{.33} & $.650\scalebox{0.75}{$\pm$}$\scalebox{0.75}{.23} & $.559\scalebox{0.75}{$\pm$}$\scalebox{0.75}{.34} & $.624$ \\
  \textcolor{blue}{\scalebox{0.8}{$\spadesuit$}}~Co-mixup~\cite{Co-mixup} & $.356\scalebox{0.75}{$\pm$}$\scalebox{0.75}{.08} & $.343\scalebox{0.75}{$\pm$}$\scalebox{0.75}{.07} & $.053\scalebox{0.75}{$\pm$}$\scalebox{0.75}{.02} & $.251$ & $.622\scalebox{0.75}{$\pm$}$\scalebox{0.75}{.30} & $.621\scalebox{0.75}{$\pm$}$\scalebox{0.75}{.21} & $.702\scalebox{0.75}{$\pm$}$\scalebox{0.75}{.21} & $.648$ \\
  \textcolor{blue}{\scalebox{0.8}{$\spadesuit$}}~MixUp~\cite{MixUp} & $.610\scalebox{0.75}{$\pm$}$\scalebox{0.75}{.14} & $.463\scalebox{0.75}{$\pm$}$\scalebox{0.75}{.15} & $.378\scalebox{0.75}{$\pm$}$\scalebox{0.75}{.15} & $.484$ & $.803\scalebox{0.75}{$\pm$}$\scalebox{0.75}{.18} & $.753\scalebox{0.75}{$\pm$}$\scalebox{0.75}{.12} & $.767\scalebox{0.75}{$\pm$}$\scalebox{0.75}{.23} & $.774$ \\
  \textcolor{blue}{\scalebox{0.8}{$\spadesuit$}}~Cutout~\cite{Cutout} & $.459\scalebox{0.75}{$\pm$}$\scalebox{0.75}{.08} & $.641\scalebox{0.75}{$\pm$}$\scalebox{0.75}{.13} & $.697\scalebox{0.75}{$\pm$}$\scalebox{0.75}{.15} & $.599$ & $.832\scalebox{0.75}{$\pm$}$\scalebox{0.75}{.17} & $.754\scalebox{0.75}{$\pm$}$\scalebox{0.75}{.11} & $.812\scalebox{0.75}{$\pm$}$\scalebox{0.75}{.13} & $.800$ \\
  \textcolor{blue}{\scalebox{0.8}{$\spadesuit$}}~CutMix~\cite{CutMix} & $.578\scalebox{0.75}{$\pm$}$\scalebox{0.75}{.06} & $.622\scalebox{0.75}{$\pm$}$\scalebox{0.75}{.12} & $.761\scalebox{0.75}{$\pm$}$\scalebox{0.75}{.11} & $.654$ & $.641\scalebox{0.75}{$\pm$}$\scalebox{0.75}{.36} & $.734\scalebox{0.75}{$\pm$}$\scalebox{0.75}{.14} & $.740\scalebox{0.75}{$\pm$}$\scalebox{0.75}{.22} & $.705$ \\
  \textcolor{blue}{\scalebox{0.8}{$\spadesuit$}}~CycleMix~\cite{CycleMix} & $.870\scalebox{0.75}{$\pm$}$\scalebox{0.75}{.06} & $.739\scalebox{0.75}{$\pm$}$\scalebox{0.75}{.05} & $.791\scalebox{0.75}{$\pm$}$\scalebox{0.75}{.07} & $.800$ & $.883\scalebox{0.75}{$\pm$}$\scalebox{0.75}{.10} & $.798\scalebox{0.75}{$\pm$}$\scalebox{0.75}{.08} & $.863\scalebox{0.75}{$\pm$}$\scalebox{0.75}{.07} & $.848$ \\
  \textcolor{blue}{\scalebox{0.8}{$\spadesuit$}}~DMPLS~\cite{DMPLS} & $.898\scalebox{0.75}{$\pm$}$\scalebox{0.75}{.08} & $.791\scalebox{0.75}{$\pm$}$\scalebox{0.75}{.12} & $.798\scalebox{0.75}{$\pm$}$\scalebox{0.75}{.19} & $.829$ & $.861\scalebox{0.75}{$\pm$}$\scalebox{0.75}{.10} & $.842\scalebox{0.75}{$\pm$}$\scalebox{0.75}{.05} & $\mathbf{.913}\scalebox{0.75}{$\pm$}$\scalebox{0.75}{.08} & $.872$ \\
  \textcolor{blue}{\scalebox{0.8}{$\spadesuit$}}~ShapePU~\cite{ShapePU} & $.919\scalebox{0.75}{$\pm$}$\scalebox{0.75}{.03} & $.832\scalebox{0.75}{$\pm$}$\scalebox{0.75}{.04} & $.804\scalebox{0.75}{$\pm$}$\scalebox{0.75}{.12} & $.852$ & $.860\scalebox{0.75}{$\pm$}$\scalebox{0.75}{.12} & $.791\scalebox{0.75}{$\pm$}$\scalebox{0.75}{.09} & $.852\scalebox{0.75}{$\pm$}$\scalebox{0.75}{.10} & $.834$ \\
  \textcolor{blue}{\scalebox{0.8}{$\spadesuit$}}~PacingPseudo~\cite{PacingPseudo} & $.734\scalebox{0.75}{$\pm$}$\scalebox{0.75}{.10} & $.793\scalebox{0.75}{$\pm$}$\scalebox{0.75}{.04} & $.890\scalebox{0.75}{$\pm$}$\scalebox{0.75}{.05} & $.806$ & $.777\scalebox{0.75}{$\pm$}$\scalebox{0.75}{.29} & $.825\scalebox{0.75}{$\pm$}$\scalebox{0.75}{.17} & $.884\scalebox{0.75}{$\pm$}$\scalebox{0.75}{.19} & $.829$ \\
  \textcolor{blue}{\scalebox{0.8}{$\spadesuit$}}~ScribFormer~\cite{ScribFormer} & $.894\scalebox{0.75}{$\pm$}$\scalebox{0.75}{.06} & $.787\scalebox{0.75}{$\pm$}$\scalebox{0.75}{.06} & $.838\scalebox{0.75}{$\pm$}$\scalebox{0.75}{.09} & $.839$ & $.903\scalebox{0.75}{$\pm$}$\scalebox{0.75}{.06} & $.848\scalebox{0.75}{$\pm$}$\scalebox{0.75}{.04} & $.845\scalebox{0.75}{$\pm$}$\scalebox{0.75}{.10} & $.865$ \\
  \textcolor{blue}{\scalebox{0.8}{$\spadesuit$}}~AIL~\cite{AIL} & $.923\scalebox{0.75}{$\pm$}$\scalebox{0.75}{.07} & $.855\scalebox{0.75}{$\pm$}$\scalebox{0.75}{.05} & $.860\scalebox{0.75}{$\pm$}$\scalebox{0.75}{.09} & $.879$ & $.905\scalebox{0.75}{$\pm$}$\scalebox{0.75}{.05} & $.809\scalebox{0.75}{$\pm$}$\scalebox{0.75}{.06} & $.845\scalebox{0.75}{$\pm$}$\scalebox{0.75}{.10} & $.853$ \\
  \textcolor{blue}{\scalebox{0.8}{$\spadesuit$}}~QMaxViT-Unet+~\cite{QMaxViT-Unet+} & $.929\scalebox{0.75}{$\pm$}$\scalebox{0.75}{.03} & $.842\scalebox{0.75}{$\pm$}$\scalebox{0.75}{.04} & $.870\scalebox{0.75}{$\pm$}$\scalebox{0.75}{.04} & $.880$ & $.919\scalebox{0.75}{$\pm$}$\scalebox{0.75}{.06} & $.868\scalebox{0.75}{$\pm$}$\scalebox{0.75}{.03} & $.845\scalebox{0.75}{$\pm$}$\scalebox{0.75}{.10} & $.876$ \\
  \textcolor{blue}{\scalebox{0.8}{$\spadesuit$}}~\textbf{Ours} & $\mathbf{.933}\scalebox{0.75}{$\pm$}$\scalebox{0.75}{.03} & $\mathbf{.859}\scalebox{0.75}{$\pm$}$\scalebox{0.75}{.03} & $\mathbf{.881}\scalebox{0.75}{$\pm$}$\scalebox{0.75}{.05} & $\mathbf{.891}$ & $\mathbf{.937}\scalebox{0.75}{$\pm$}$\scalebox{0.75}{.04} & $\mathbf{.904}\scalebox{0.75}{$\pm$}$\scalebox{0.75}{.02} & $.892\scalebox{0.75}{$\pm$}$\scalebox{0.75}{.05} & $\mathbf{.911}$ \\
  \midrule
  \textcolor{red}{\scalebox{0.8}{$\blacksquare$}}~UNet$_{F}$~\cite{UNetF} & $.890\scalebox{0.75}{$\pm$}$\scalebox{0.75}{.05} & $.726\scalebox{0.75}{$\pm$}$\scalebox{0.75}{.10} & $.688\scalebox{0.75}{$\pm$}$\scalebox{0.75}{.18} & $.768$ & $.898\scalebox{0.75}{$\pm$}$\scalebox{0.75}{.09} & $.839\scalebox{0.75}{$\pm$}$\scalebox{0.75}{.04} & $.804\scalebox{0.75}{$\pm$}$\scalebox{0.75}{.12} & $.847$ \\
  \textcolor{red}{\scalebox{0.8}{$\blacksquare$}}~UNet\scalebox{0.8}{+}$_{F}$~\cite{CycleMix} & $.857\scalebox{0.75}{$\pm$}$\scalebox{0.75}{.06} & $.720\scalebox{0.75}{$\pm$}$\scalebox{0.75}{.08} & $.689\scalebox{0.75}{$\pm$}$\scalebox{0.75}{.12} & $.755$ & $.883\scalebox{0.75}{$\pm$}$\scalebox{0.75}{.13} & $.831\scalebox{0.75}{$\pm$}$\scalebox{0.75}{.09} & $.870\scalebox{0.75}{$\pm$}$\scalebox{0.75}{.10} & $.862$ \\
  \textcolor{red}{\scalebox{0.8}{$\blacksquare$}}~UNet\scalebox{0.8}{++}$_{F}$~\cite{UNet++F} & $.847\scalebox{0.75}{$\pm$}$\scalebox{0.75}{.09} & $.678\scalebox{0.75}{$\pm$}$\scalebox{0.75}{.11} & $.717\scalebox{0.75}{$\pm$}$\scalebox{0.75}{.08} & $.756$ & $.889\scalebox{0.75}{$\pm$}$\scalebox{0.75}{.09} & $.820\scalebox{0.75}{$\pm$}$\scalebox{0.75}{.04} & $.733\scalebox{0.75}{$\pm$}$\scalebox{0.75}{.18} & $.814$ \\
  \textcolor{red}{\scalebox{0.8}{$\blacksquare$}}~Puzzle Mix$_{F}$~\cite{PuzzleMix} & $.867\scalebox{0.75}{$\pm$}$\scalebox{0.75}{.04} & $.742\scalebox{0.75}{$\pm$}$\scalebox{0.75}{.04} & $.759\scalebox{0.75}{$\pm$}$\scalebox{0.75}{.04} & $.789$ & $.912\scalebox{0.75}{$\pm$}$\scalebox{0.75}{.08} & $.842\scalebox{0.75}{$\pm$}$\scalebox{0.75}{.08} & $.887\scalebox{0.75}{$\pm$}$\scalebox{0.75}{.07} & $.880$ \\
  \textcolor{red}{\scalebox{0.8}{$\blacksquare$}}~CycleMix$_{F}$~\cite{CycleMix} & $.864\scalebox{0.75}{$\pm$}$\scalebox{0.75}{.03} & $.785\scalebox{0.75}{$\pm$}$\scalebox{0.75}{.04} & $.781\scalebox{0.75}{$\pm$}$\scalebox{0.75}{.07} & $.810$ & $.919\scalebox{0.75}{$\pm$}$\scalebox{0.75}{.07} & $.858\scalebox{0.75}{$\pm$}$\scalebox{0.75}{.06} & $.882\scalebox{0.75}{$\pm$}$\scalebox{0.75}{.09} & $.886$ \\
  \textcolor{red}{\scalebox{0.8}{$\blacksquare$}}~nnU-Net~\cite{nnUNet} & $.940\scalebox{0.75}{$\pm$}$\scalebox{0.75}{.04} & $.865\scalebox{0.75}{$\pm$}$\scalebox{0.75}{.03} & $.886\scalebox{0.75}{$\pm$}$\scalebox{0.75}{.04} & $.897$  & $.939\scalebox{0.75}{$\pm$}$\scalebox{0.75}{.05} & $.918\scalebox{0.75}{$\pm$}$\scalebox{0.75}{.02} & $.900\scalebox{0.75}{$\pm$}$\scalebox{0.75}{.03} & $.917$ \\
  \bottomrule
  \end{tabular}
  \label{tab:performance_comparison}
  \endgroup
\end{table*}

\section{Experiments}

\subsection{Datasets and Evaluation Metrics}We evaluated our proposed method on two publicly available medical image segmentation datasets: MSCMRseg \cite{MSCMR} and ACDC \cite{ACDC}. Although these datasets are widely used, we adhered to the original training and testing splits. The MSCMRseg dataset contains late gadolinium enhancement cardiac MR images from 45 patients with cardiomyopathy. It provides pixel-level annotations for three cardiac structures—right ventricle (RV), left ventricle (LV) and myocardium (Myo)—with both gold standard and scribble-style annotations. In our experiments, we strictly followed the data split method proposed by Zhuang et al. \cite{CycleMix}, using 25 cases for training, 5 cases for validation, and 15 cases for testing. Similarly, the ACDC dataset consists of cine-MR images from 100 patients, acquired using 1.5T and 3T MRI scanners with varying temporal resolutions. It provides annotations for each patient at end-diastolic and end-systolic phases, covering RV, LV, and Myo. We also followed the standard data split strategy reported in Zhuang et al. \cite{CycleMix}, selecting 35 cases for training, 15 for validation, and 15 for testing, while excluding the remaining 35 cases to ensure fair comparison with previous works. 
\subsection{Experimental Details}
All experiments were conducted using the PyTorch framework and accelerated on an NVIDIA RTX 4090 GPU (24GB). During preprocessing,  each image was standardized to have zero mean and unit variance. All images were then resized to 224 × 224 pixels to match the network input requirements. The Adam optimizer was used to minimize the loss function defined in Eq.\ref{eq_11}, with an initial learning rate of 0.0001 and an exponential decay rate of 0.95. Training was performed over 1000 epochs. The hyperparameters in Eq.\ref{eq_11} were empirically set to $\lambda_1$ = 1.0, $\lambda_2$ = 0.3 and $\lambda_3$ = 0.1. During inference, the model’s predictions on the original input images were directly adopted as the final segmentation results, without any post-processing or additional refinement. The segmentation performance was evaluated using the Dice coefficient. 

\begin{figure*}[t] 
  \centering
  \includegraphics[width=\textwidth]{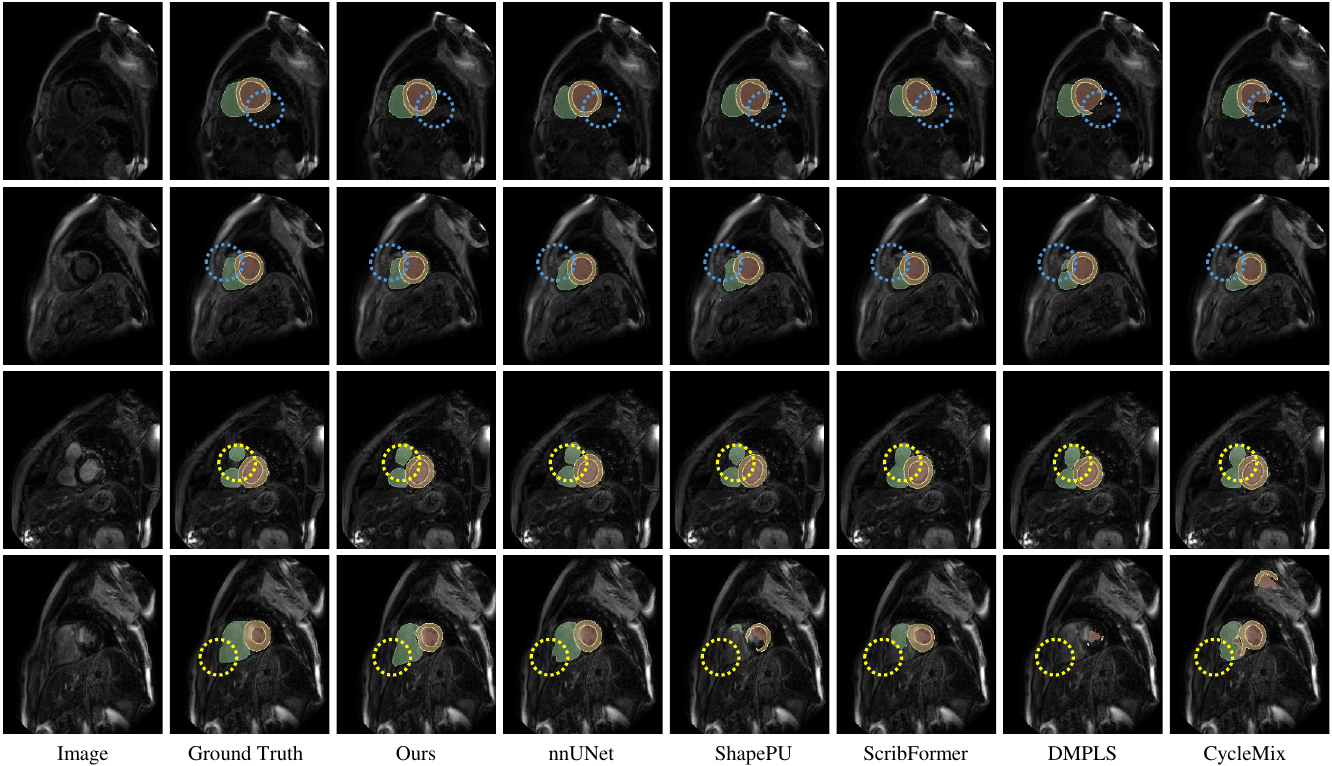} 
  \caption{Visualization on four typical cases from the MSCMRseg dataset for illustration and comparison. Blue dashed circles highlight regions with ambiguous boundaries, and yellow dashed circles indicate areas of over- or under-segmentation.}
  \label{fig:fig2}
\end{figure*}

\subsection{Results on MSCMRseg and ACDC Datasets}Table.\ref{tab:performance_comparison} presents a comparative analysis of our method’s performance against other methods on the MSCMRseg and ACDC datasets. Two types of supervision were evaluated: scribble supervision(blue spade \textcolor{blue}{$\spadesuit$}) and full supervision(red square \textcolor{red}{$\blacksquare$}). For the scribble-supervised category, we include UNet+ \cite{CycleMix} as a baseline, along with various augmentation strategies applied to UNet+, such as Puzzle Mix \cite{PuzzleMix}, Co-mixup \cite{Co-mixup}, Mixup \cite{MixUp}, Cutout \cite{Cutout}, CutMix \cite{CutMix}, CycleMix \cite{CycleMix} and ShapePU \cite{ShapePU}. Additionally, we evaluate other scribble-based segmentation frameworks that incorporate pseudo-labeling mechanisms, including DMPLS \cite{DMPLS}, ScribFormer \cite{ScribFormer}, PacingPseudo \cite{PacingPseudo}, AIL \cite{AIL} and QMaxViT-Unet+ \cite{QMaxViT-Unet+}. In the fully supervised category, we compare several UNet-based models, such as UNet$_{F}$ \cite{UNetF}, UNet+$_{F}$ \cite{CycleMix}, UNet++$_{F}$ \cite{UNet++F}, and nnUNet \cite{nnUNet}, as well as the fully supervised versions of weakly supervised frameworks. All results are reported as mean ± standard deviation. 

As shown in Table.\ref{tab:performance_comparison}, our method achieves a superior average Dice score of 89.1\% on the MSCMRseg dataset, outperforming all other weakly supervised methods. It surpasses the previous state-of-the-art method, QMaxViT-Unet+, by 1.1\%. Compared with CycleMix, which also employs augmentation strategies, our method improves the average Dice score by 9.1\% (89.1\% vs. 80.0\%). In comparison to pseudo-label-based methods such as DMPLS and AIL, our method yields gains of 6.2\% (89.1\% vs. 82.9\%) and 1.2\% (89.1\% vs. 87.9\%), respectively. Notably, our method achieves performance comparable to the best fully supervised method (nnUNet), with a Dice score of 89.1\% versus 89.7\%.

On the ACDC dataset, our method also achieves an average Dice score of 91.1\%, significantly outperforming other weakly supervised methods. It improves upon the previous best-performing method, QMaxViT-Unet+, by 3.5\%. Compared with CycleMix, TABNet achieves a 6.3\% higher Dice score (91.1\% vs. 84.8\%). When compared to other pseudo-label-based methods, our method shows improvements of 3.9\% over DMPLS (91.1\% vs. 87.2\%) and 5.8\% over AIL (91.1\% vs. 85.3\%). Remarkably, our performance is close to that of the state-of-the-art fully supervised method nnUNet (91.1\% vs. 91.7\%).

\begin{figure*}[t] 
  \centering
  \includegraphics[width=\textwidth]{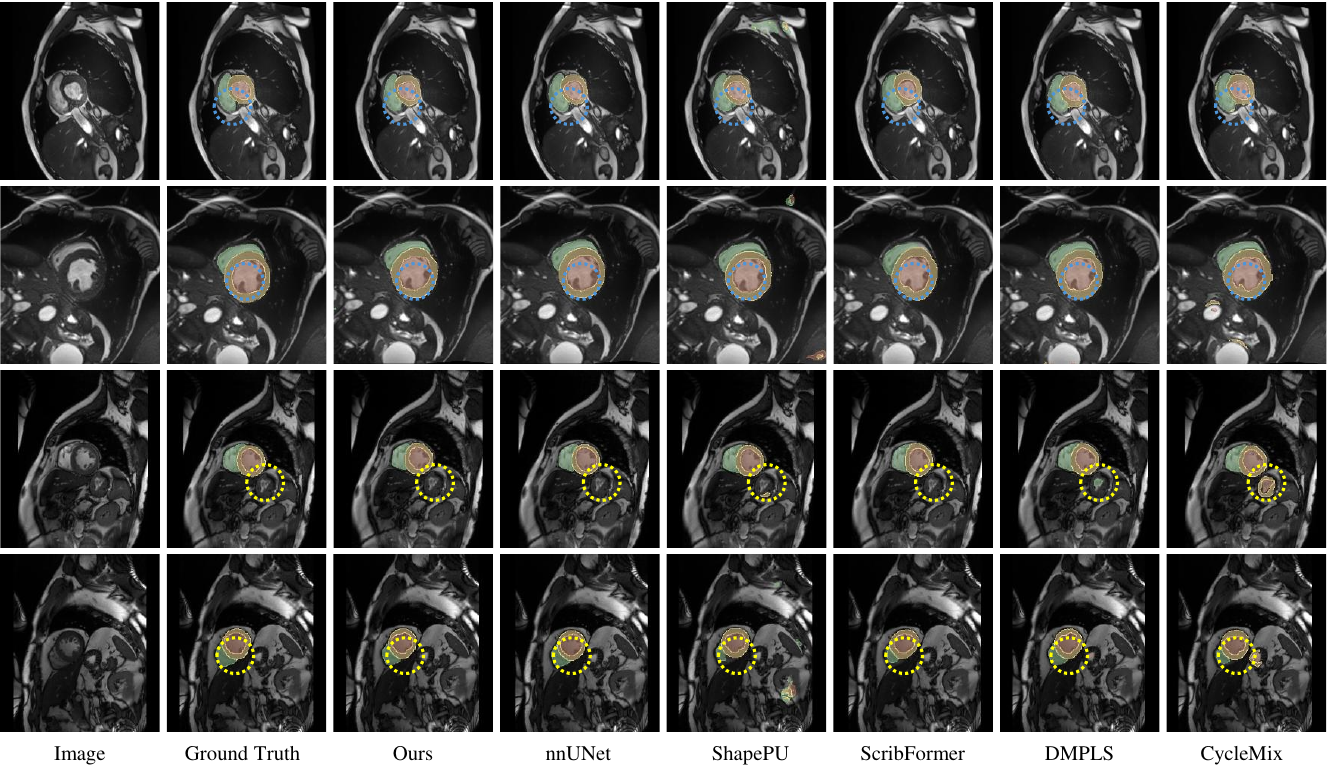} 
  \caption{Visualization on four typical cases from the ACDC dataset for illustration and comparison. Blue dashed circles highlight regions with ambiguous boundaries, and yellow dashed circles indicate areas of over- or under-segmentation.}
  \label{fig:fig3}
\end{figure*}

\textbf{Qualitative Visualization:} As shown in Fig.\ref{fig:fig2} and Fig.\ref{fig:fig3}, we present a qualitative comparison between our method and several state-of-the-art methods. The blue dashed circles highlight regions with ambiguous boundaries, where existing weakly supervised methods often produce disconnected or imprecise contours. In contrast, our method yields more accurate and coherent boundary predictions, closely aligning with the ground truth. This improvement primarily benefits from our proposed pseudo-label fusion mechanism, which dynamically integrates predictions from the intensity and jigsaw branches using a loss-weighted strategy. The fused pseudo-labels provide more reliable supervision, especially in boundary regions. Building upon this, the boundary-aware pseudo-label supervision module further refines contour precision by extracting soft boundary maps via min-pooling and enforcing consistency between multi-branch predictions and pseudo-label boundaries. 

In addition, the yellow dashed circles indicate regions prone to over-segmentation or under-segmentation, which typically arise due to incomplete semantic representations or limited contextual understanding. Our method effectively mitigates these issues, producing more complete and structurally coherent segmentation results. This can be attributed to the proposed TAS module, which incorporates intensity, cutout, and jigsaw perturbations during training. Intensity perturbation simulates grayscale variations to encourage the network to learn texture-level features; cutout augmentation occludes organ regions, prompting the model to focus on local anatomical structures; and jigsaw augmentation disrupts spatial continuity, guiding the model to understand global spatial layouts and organ configurations. As a result, the model is capable of recovering missing structures and enhancing region-level integrity under sparse supervision.These qualitative results demonstrate that our method not only resolves boundary ambiguity and improves region completeness under sparse supervision, but also achieves segmentation quality comparable to fully supervised counterparts. 

\begin{table}[!htbp]
  \centering
  \small
  \caption{Ablation studies on the MSCMRseg dataset under different settings. Symbol $^\ast$ indicates statistically significant improvement ($p < 0.05$). Bold denotes the best performance. }
  \label{tab:ablation_studies}
  \renewcommand{\arraystretch}{1.2}
  \setlength{\tabcolsep}{4pt} 
  \begingroup
  \fontfamily{ptm}\selectfont  
  \begin{tabular*}{\columnwidth}{@{\extracolsep{\fill}}ccc|ccc|l} 
    \toprule
    $\mathcal{L}_{TAS}$ & $\mathcal{L}_{PL}$ & $\mathcal{L}_{BD}$ & LV & Myo & RV & \multicolumn{1}{l}{Avg} \\
    \midrule
    $\times$ & $\times$ & $\times$ & $.894\scalebox{0.75}{$\pm$}$\scalebox{0.75}{.04} & $.742\scalebox{0.75}{$\pm$}$\scalebox{0.75}{.06} & $.821\scalebox{0.75}{$\pm$}$\scalebox{0.75}{.08} & $.819$ \\
    $\checkmark$ & $\times$ & $\times$ & $.930\scalebox{0.75}{$\pm$}$\scalebox{0.75}{.02} & $.838\scalebox{0.75}{$\pm$}$\scalebox{0.75}{.05} & $.734\scalebox{0.75}{$\pm$}$\scalebox{0.75}{.11} & $.834^\ast$ \\
    $\checkmark$ & $\checkmark$ & $\times$ & $.929\scalebox{0.75}{$\pm$}$\scalebox{0.75}{.03} & $.844\scalebox{0.75}{$\pm$}$\scalebox{0.75}{.03} & $.861\scalebox{0.75}{$\pm$}$\scalebox{0.75}{.05} & $.878^\ast$ \\
    $\checkmark$ & $\checkmark$ & $\checkmark$ & $\mathbf{.933}\scalebox{0.75}{$\pm$}$\scalebox{0.75}{.03} & $\mathbf{.859}\scalebox{0.75}{$\pm$}$\scalebox{0.75}{.03} & $\mathbf{.881}\scalebox{0.75}{$\pm$}$\scalebox{0.75}{.05} & $\mathbf{.891}^\ast$ \\
    \bottomrule
  \end{tabular*}
  \endgroup
\end{table}

\subsection{Ablation Study}

\subsubsection{Effectiveness of Loss Function}

We conducted an ablation study to evaluate the contribution of each module in our method, using the MSCMRseg dataset. The results are shown in Table.\ref{tab:ablation_studies}. When no strategy was applied, the model achieved a Dice score of 81.9\%. The introduction of the TAS module improved the average Dice score by 1.5\%, demonstrating the effectiveness of this module in feature learning. With the addition of PL, the Dice score increased further by 4.4\%. Finally, the incorporation of boundary-aware loss $\mathcal{L}_{BD}$ significantly enhanced the boundary segmentation accuracy, achieving an average Dice score of 89.1\%, surpassing all other configurations. This highlights the advantage of boundary supervision in refining boundary segmentation. 

\begin{table}[htbp]
    \centering
    \small
    \caption{An ablation study on the number of branches in the TAS module was conducted on the MSCMRseg dataset. Bold values indicate the best performance. }
    \label{tab:ablation_TAS}
    \renewcommand{\arraystretch}{1.2}
    \setlength{\tabcolsep}{1pt}  
    \begingroup
    \fontfamily{ptm}\selectfont  
    
    \begin{tabular*}{\columnwidth}{@{\extracolsep{\fill}}c|ccc|c}  
    \toprule
    \multicolumn{1}{c|}{Data augmentation} & \multicolumn{1}{c}{LV} & \multicolumn{1}{c}{Myo} & \multicolumn{1}{c|}{RV} & \multicolumn{1}{c}{Avg} \\
    \midrule
    Cutout  & $.914\scalebox{0.75}{$\pm$}$\scalebox{0.75}{.04} & $.827\scalebox{0.75}{$\pm$}$\scalebox{0.75}{.05} & $.842\scalebox{0.75}{$\pm$}$\scalebox{0.75}{.08} & $.861$ \\
    Jigsaw  & $.925\scalebox{0.75}{$\pm$}$\scalebox{0.75}{.03} & $.826\scalebox{0.75}{$\pm$}$\scalebox{0.75}{.04} & $.855\scalebox{0.75}{$\pm$}$\scalebox{0.75}{.06} & $.869$ \\
    Intensity   & $.914\scalebox{0.75}{$\pm$}$\scalebox{0.75}{.03} & $.826\scalebox{0.75}{$\pm$}$\scalebox{0.75}{.04} & $.854\scalebox{0.75}{$\pm$}$\scalebox{0.75}{.07} & $.865$ \\
    \midrule
    (Cutout, Jigsaw)  & $.932\scalebox{0.75}{$\pm$}$\scalebox{0.75}{.03} & $.847\scalebox{0.75}{$\pm$}$\scalebox{0.75}{.03} & $.869\scalebox{0.75}{$\pm$}$\scalebox{0.75}{.05} & $.883$ \\
    (Cutout, Intensity)   & $.930\scalebox{0.75}{$\pm$}$\scalebox{0.75}{.03} & $.844\scalebox{0.75}{$\pm$}$\scalebox{0.75}{.03} & $.868\scalebox{0.75}{$\pm$}$\scalebox{0.75}{.06} & $.881$ \\
    (Jigsaw, Intensity)   & $.931\scalebox{0.75}{$\pm$}$\scalebox{0.75}{.03} & $.845\scalebox{0.75}{$\pm$}$\scalebox{0.75}{.03} & $.878\scalebox{0.75}{$\pm$}$\scalebox{0.75}{.05} & $.885$ \\
    \midrule
    (Intensity, Cutout, Jigsaw)  & $\mathbf{.933}\scalebox{0.75}{$\pm$}$\scalebox{0.75}{.03} & $\mathbf{.859}\scalebox{0.75}{$\pm$}$\scalebox{0.75}{.03} & $\mathbf{.881}\scalebox{0.75}{$\pm$}$\scalebox{0.75}{.05} & $\mathbf{.891}$ \\
    \bottomrule
    \end{tabular*}
    \endgroup
\end{table}

\subsubsection{Effectiveness of TAS Module}

As shown in Table.\ref{tab:ablation_TAS}, we conducted an ablation study on the MSCMRseg dataset to evaluate the effectiveness of the TAS module by testing each of the three augmentation branches—intensity, cropping, and jigsaw—individually, in pairs, and in full combination. In both the single- and dual-branch experiments, the remaining branches were fed with the original (unaugmented) image to ensure a fair comparison and isolate the contributions of the selected augmentations. Each augmentation branch individually contributed to performance gains. Notably, the jigsaw branch achieved the best single-branch performance with an average Dice score of 86.9\%, outperforming Intensity and Cutout branches (both 86.5\% and 86.1\%, respectively). Specifically, Jigsaw improved LV segmentation from 91.4\% to 92.5\% and RV from 85.4\% to 85.5\%, suggesting its strength in preserving structural layout. Combining two branches further enhanced segmentation performance. The best dual-branch combination, Jigsaw and Intensity, achieved an average Dice score of 88.5\%, representing a 1.6\% improvement over the best single-branch (jigsaw). This setting improved the RV score to 87.8\%, showing clear advantages in boundary preservation and anatomical separation. Ultimately, integrating all three augmentations—intensity, cutout, and Jigsaw—yielded the best overall results, with an average Dice of 0.891, further improving performance by 0.6\% over the best two-branch setting and 2.6\% over the single Intensity branch. This full configuration achieved the highest scores in all categories, with LV: 93.3\%, Myo: 85.9\%, and RV: 88.1\%, validating the complementary nature of texture-level (intensity), local structural (cutout), and global contextual (jigsaw) information for robust and generalizable feature learning under weak supervision. 

\begin{table}[t]
    \centering
    \small
    \caption{Ablation of the number of fusion branches for the PL module on the MSCMRseg dataset, and comparison of PL generation methods. Bold indicates the best performance.}
    \label{tab:ablation_PL}
    \renewcommand{\arraystretch}{1.2}
    \setlength{\tabcolsep}{1pt}
    \begingroup
    \fontfamily{ptm}\selectfont  
    \begin{tabular*}{\columnwidth}{@{\extracolsep{\fill}}c|ccc|c}
    \toprule
    \multicolumn{1}{c|}{PL Generation Method} & \multicolumn{1}{c}{LV} & \multicolumn{1}{c}{Myo} & \multicolumn{1}{c|}{RV} & \multicolumn{1}{c}{Avg} \\
    \midrule
    $y_i$ & $.924\scalebox{0.75}{$\pm$}$\scalebox{0.75}{.04} & $.845\scalebox{0.75}{$\pm$}$\scalebox{0.75}{.03} & $.863\scalebox{0.75}{$\pm$}$\scalebox{0.75}{.06} & $.878$ \\
    $y_j$ & $.931\scalebox{0.75}{$\pm$}$\scalebox{0.75}{.03} & $.842\scalebox{0.75}{$\pm$}$\scalebox{0.75}{.03} & $.868\scalebox{0.75}{$\pm$}$\scalebox{0.75}{.05} & $.880$ \\
    $y_k$ & $.932\scalebox{0.75}{$\pm$}$\scalebox{0.75}{.03} & $.840\scalebox{0.75}{$\pm$}$\scalebox{0.75}{.04} & $.874\scalebox{0.75}{$\pm$}$\scalebox{0.75}{.06} & $.882$ \\
    \midrule
    PL($y_i$,$y_j$) & $.929\scalebox{0.75}{$\pm$}$\scalebox{0.75}{.03} & $.843\scalebox{0.75}{$\pm$}$\scalebox{0.75}{.04} & $.880\scalebox{0.75}{$\pm$}$\scalebox{0.75}{.05} & $.884$ \\
    PL($y_i$,$y_k$) & $.933\scalebox{0.75}{$\pm$}$\scalebox{0.75}{.03} & $.843\scalebox{0.75}{$\pm$}$\scalebox{0.75}{.04} & $.876\scalebox{0.75}{$\pm$}$\scalebox{0.75}{.05} & $.884$ \\
    PL($y_j$,$y_k$) & $\mathbf{.933}\scalebox{0.75}{$\pm$}$\scalebox{0.75}{.03} & $\mathbf{.859}\scalebox{0.75}{$\pm$}$\scalebox{0.75}{.03} & $\mathbf{.881}\scalebox{0.75}{$\pm$}$\scalebox{0.75}{.05} & $\mathbf{.891}$ \\
    \midrule
    PL($y_i$,$y_j$,$y_k$)  & $.932\scalebox{0.75}{$\pm$}$\scalebox{0.75}{.03} & $.842\scalebox{0.75}{$\pm$}$\scalebox{0.75}{.04} & $.880\scalebox{0.75}{$\pm$}$\scalebox{0.75}{.05} & $.885$ \\
    \midrule
    PL($y_j$,$y_k$)& $\mathbf{.933}\scalebox{0.75}{$\pm$}$\scalebox{0.75}{.03} & $\mathbf{.859}\scalebox{0.75}{$\pm$}$\scalebox{0.75}{.03} & $\mathbf{.881}\scalebox{0.75}{$\pm$}$\scalebox{0.75}{.05} & $\mathbf{.891}$ \\
    Average($y_j$,$y_k$) & $.929\scalebox{0.75}{$\pm$}$\scalebox{0.75}{.03} & $.842\scalebox{0.75}{$\pm$}$\scalebox{0.75}{.04} & $.878\scalebox{0.75}{$\pm$}$\scalebox{0.75}{.06} & $.883$ \\
    Random($y_j$,$y_k$) & $.931\scalebox{0.75}{$\pm$}$\scalebox{0.75}{.03} & $.845\scalebox{0.75}{$\pm$}$\scalebox{0.75}{.03} & $.876\scalebox{0.75}{$\pm$}$\scalebox{0.75}{.04} & $.884$ \\
    \bottomrule
    \end{tabular*}
    \endgroup
\end{table}

\subsubsection{Effectiveness of PL Module}

As shown in Table~\ref{tab:ablation_PL}, we conducted a comprehensive evaluation of the PL module by testing three pseudo-label generation branches: cropping augmentation ($y_i$), jigsaw augmentation ($y_j$), and intensity augmentation ($y_k$), in various combinations. When applying each augmentation individually for pseudo-label generation, the intensity augmentation branch ($y_k$) achieved the highest average Dice score of 88.2\%, showing particularly strong performance on LV segmentation (93.2\%). In RV segmentation, intensity augmentation ($y_k$) also outperformed jigsaw augmentation ($y_j$) with scores of 87.4\% versus 86.8\%. The optimal dual-branch combination was jigsaw and intensity augmentation (PL($y_j$, $y_k$)), which achieved an average Dice score of 89.1\% and significant improvement in myocardial segmentation (85.9\%), representing gains of 1.7\% and 1.9\% over single $y_j$ and $y_k$ branches, respectively. This combination also maintained excellent LV performance at 93.3\%. In contrast, cropping-based combinations (PL($y_i$, $y_j$) and PL($y_i$, $y_k$)) both achieved 88.4\%. The full three-branch fusion (PL($y_i$, $y_j$, $y_k$)) achieved an average Dice of 88.5\%, which slightly outperformed individual branches but remained 0.6\% lower than the best dual-branch combination, with RV segmentation at 88.0\% compared to 88.1\% for PL($y_j$, $y_k$). A comparison of fusion strategies revealed that loss-weighted fusion (89.1\%) consistently outperformed both average weighting (88.3\%) and random weighting (88.4\%), with the most significant improvement seen in myocardial segmentation (85.9\% vs. 84.2\% for average weighting). These findings demonstrate that jigsaw and intensity augmentations offer the most effective complementarity, while cropping provides limited additional benefit. Moreover, adaptive loss-weighting yields consistent advantages over fixed fusion strategies in weakly supervised medical image segmentation.

\begin{table}[t]
    \centering
    \small
    \caption{Performance under different $\lambda$ values with fixed other perturbation strengths.}
    \label{tab:ablation_LAMUDA}
    \renewcommand{\arraystretch}{1.2}
    \setlength{\tabcolsep}{6pt} 
    \begingroup
    \fontfamily{ptm}\selectfont  
    
    \begin{tabular}{l|ccc|c}
    \toprule
    $\lambda$ & \multicolumn{1}{c}{LV} & \multicolumn{1}{c}{Myo} & \multicolumn{1}{c|}{RV} & \multicolumn{1}{c}{Avg} \\
    \midrule
    $\lambda_1$ & \multicolumn{4}{l}{$\lambda_2=0.3,\ \lambda_3=0.1$} \\
    \midrule
    1.0 & $\mathbf{.933}\scalebox{0.75}{$\pm$}$\scalebox{0.75}{.03} & $\mathbf{.859}\scalebox{0.75}{$\pm$}$\scalebox{0.75}{.03} & $\mathbf{.881}\scalebox{0.75}{$\pm$}$\scalebox{0.75}{.05} & $\mathbf{.891}$ \\
    0.8 & $.932\scalebox{0.75}{$\pm$}$\scalebox{0.75}{.03} & $.858\scalebox{0.75}{$\pm$}$\scalebox{0.75}{.03} & $.877\scalebox{0.75}{$\pm$}$\scalebox{0.75}{.06} & $.889$ \\
    0.5 & $.927\scalebox{0.75}{$\pm$}$\scalebox{0.75}{.03} & $.841\scalebox{0.75}{$\pm$}$\scalebox{0.75}{.03} & $.869\scalebox{0.75}{$\pm$}$\scalebox{0.75}{.06} & $.879$ \\
    0.3 & $.926\scalebox{0.75}{$\pm$}$\scalebox{0.75}{.03} & $.837\scalebox{0.75}{$\pm$}$\scalebox{0.75}{.04} & $.863\scalebox{0.75}{$\pm$}$\scalebox{0.75}{.06} & $.875$ \\
    0.1 & $.922\scalebox{0.75}{$\pm$}$\scalebox{0.75}{.03} & $.833\scalebox{0.75}{$\pm$}$\scalebox{0.75}{.04} & $.840\scalebox{0.75}{$\pm$}$\scalebox{0.75}{.07} & $.865$ \\
    \midrule
    $\lambda_2$ & \multicolumn{4}{l}{$\lambda_1=1.0,\ \lambda_3=0.1$} \\
    \midrule
    1.0 & $.927\scalebox{0.75}{$\pm$}$\scalebox{0.75}{.04} & $.836\scalebox{0.75}{$\pm$}$\scalebox{0.75}{.04} & $.866\scalebox{0.75}{$\pm$}$\scalebox{0.75}{.06} & $.876$ \\
    0.8 & $.933\scalebox{0.75}{$\pm$}$\scalebox{0.75}{.03} & $.841\scalebox{0.75}{$\pm$}$\scalebox{0.75}{.04} & $.860\scalebox{0.75}{$\pm$}$\scalebox{0.75}{.06} & $.878$ \\
    0.5 & $.933\scalebox{0.75}{$\pm$}$\scalebox{0.75}{.03} & $.843\scalebox{0.75}{$\pm$}$\scalebox{0.75}{.03} & $.872\scalebox{0.75}{$\pm$}$\scalebox{0.75}{.05} & $.883$ \\
    0.3 & $\mathbf{.933}\scalebox{0.75}{$\pm$}$\scalebox{0.75}{.03} & $\mathbf{.859}\scalebox{0.75}{$\pm$}$\scalebox{0.75}{.03} & $\mathbf{.881}\scalebox{0.75}{$\pm$}$\scalebox{0.75}{.05} & $\mathbf{.891}$ \\
    0.1 & $.927\scalebox{0.75}{$\pm$}$\scalebox{0.75}{.03} & $.840\scalebox{0.75}{$\pm$}$\scalebox{0.75}{.04} & $.868\scalebox{0.75}{$\pm$}$\scalebox{0.75}{.06} & $.878$ \\
    \midrule
    $\lambda_3$ & \multicolumn{4}{l}{$\lambda_1=1.0,\ \lambda_2=0.3$} \\
    \midrule
    1.0 & $.918\scalebox{0.75}{$\pm$}$\scalebox{0.75}{.04} & $.827\scalebox{0.75}{$\pm$}$\scalebox{0.75}{.05} & $.857\scalebox{0.75}{$\pm$}$\scalebox{0.75}{.06} & $.867$ \\
    0.8 & $.928\scalebox{0.75}{$\pm$}$\scalebox{0.75}{.03} & $.841\scalebox{0.75}{$\pm$}$\scalebox{0.75}{.04} & $.863\scalebox{0.75}{$\pm$}$\scalebox{0.75}{.06} & $.877$ \\
    0.5 & $.925\scalebox{0.75}{$\pm$}$\scalebox{0.75}{.03} & $.837\scalebox{0.75}{$\pm$}$\scalebox{0.75}{.03} & $.876\scalebox{0.75}{$\pm$}$\scalebox{0.75}{.05} & $.879$ \\
    0.3 & $.932\scalebox{0.75}{$\pm$}$\scalebox{0.75}{.03} & $.844\scalebox{0.75}{$\pm$}$\scalebox{0.75}{.03} & $.868\scalebox{0.75}{$\pm$}$\scalebox{0.75}{.06} & $.882$ \\
    0.1 & $\mathbf{.933}\scalebox{0.75}{$\pm$}$\scalebox{0.75}{.03} & $\mathbf{.859}\scalebox{0.75}{$\pm$}$\scalebox{0.75}{.03} & $\mathbf{.881}\scalebox{0.75}{$\pm$}$\scalebox{0.75}{.05} & $\mathbf{.891}$ \\
    \bottomrule
    \end{tabular}
    \endgroup
\end{table}

\subsubsection{Ablation of Hyperparameters}

To investigate the impact of the weight coefficients in Eq.~\ref{eq_11} on segmentation performance, we conducted a comprehensive ablation study focusing on the hyperparameters $\lambda_1$, $\lambda_2$, and $\lambda_3$, corresponding to the TAS module, the PL module, and the boundary-aware loss, respectively. On the MSCMRseg dataset, each $\lambda$ was individually set to 1.0, 0.8, 0.5, 0.3, and 0.1, while keeping the other two fixed to evaluate its specific influence. The optimal configuration ($\lambda_1=1.0$, $\lambda_2=0.3$, $\lambda_3=0.1$) achieved the best overall performance with an average Dice score of 89.1\%. When $\lambda_2$ was reduced from 1.0 to 0.3 (with $\lambda_1=1.0$ and $\lambda_3=0.1$ fixed), the average Dice improved from 87.6\% to 89.1\%. This improvement was especially evident in myocardial segmentation, where the Dice score increased from 83.6\% to 85.9\%. LV and RV segmentation also improved, rising from 92.7\% to 93.3\%, and from 86.6\% to 88.1\%, respectively. In experiments varying $\lambda_1$ (with $\lambda_2=0.3$ and $\lambda_3=0.1$), performance declined steadily as $\lambda_1$ decreased. For instance, the average Dice dropped from 89.1\% to 86.5\% when $\lambda_1$ decreased from 1.0 to 0.1, with RV segmentation showing the largest decrease, from 88.1\% to 84.0\%. Similarly, adjusting $\lambda_3$ (fixing $\lambda_1=1.0$ and $\lambda_2=0.3$) revealed that a smaller boundary loss weight leads to better results. The Dice score increased from 86.7\% at $\lambda_3=1.0$ to 89.1\% at $\lambda_3=0.1$, with myocardial segmentation improving from 82.7\% to 85.9\%. These results indicate that: (1) the TAS module benefits from strong weighting ($\lambda_1=1.0$), particularly for RV segmentation; (2) the PL module performs best at a moderate weight ($\lambda_2=0.3$), especially enhancing Myo accuracy; and (3) a smaller boundary loss weight ($\lambda_3=0.1$) yields superior overall segmentation. The model also demonstrated stability across different configurations, with standard deviations below $\pm$7\%. 

\section{Conclusions}

We propose TABNet, an innovative weakly supervised medical image segmentation framework that integrates a TAS module and a BAP module. The TAS module enhances the model’s ability to learn rich and robust features from sparsely annotated data by introducing intensity, cutout, and jigsaw augmentations, along with a region recovery strategy. The BAP module generates reliable pseudo-labels through a loss-weighted fusion of dual-branch predictions and enforces boundary consistency via soft boundary extraction and supervision. This dual-module design enables the model to achieve both accurate region prediction and precise boundary delineation. Extensive experiments on two public datasets, ACDC and MSCMRseg, demonstrate that TABNet outperforms existing weakly supervised methods and closely methods the performance of fully supervised models. In future work, we intend to advance this framework toward more challenging tasks, such as multi-organ analysis, cross-modality adaptation, and large-scale multi-center data analysis, with the goal of enhancing its clinical applicability.

\section*{Ethics statement}

This study relies entirely on computer simulations and does not involve any experiments with human participants, animals, or biological specimens. The parameters used in the simulations are derived from publicly available data sources or previously published studies. Therefore, no ethical approval or informed consent was required. 

\section*{CRediT authorship contribution statement}
    \textbf{Peilin Zhang:} Writing - review \& editing, Writing – original draft, Methodology. \textbf{Shaouxan Wu:} Writing - review \& editing, Validation, Software. \textbf{Jun Feng:} Writing - review \& editing, Supervision. \textbf{Zhuo Jin:} Validation. \textbf{Zhizezhang Gao:} Writing - review \& editing. \textbf{Jingkun Chen:} Writing - review \& editing. \textbf{Yaqiong Xing:} Funding acquisition, Supervision. \textbf{Xiao Zhang:} Funding acquisition, Writing - review \& editing, Resources. 

\section*{Declaration of Competing Interest}

The authors declare that they have no known competing financial interests or personal relationships that could have appeared to influence the work reported in this paper. 

\section*{Acknowledgments}

This work was supported by the National Natural Science Foundation of China (No. 62403380) and Shaanxi Province Postdoctoral Science Foundation (No.2024BSHSD\\ZZ042). 

\printcredits



\end{document}